\documentclass[11pt,a4paper]{article}
\usepackage[hyperref]{acl2020}
\usepackage{times}
\usepackage{latexsym}

\usepackage{microtype}

\usepackage[utf8]{inputenc}
\usepackage{url}            
\usepackage{booktabs}       
\usepackage{amsmath}
\usepackage{amsfonts}       
\usepackage{nicefrac}       
\usepackage{natbib}
\usepackage{array}
\usepackage{makecell}
\usepackage{multirow}
\usepackage{amsfonts}
\usepackage{graphicx}
\usepackage{wrapfig} 
\usepackage{float}
\usepackage{caption} 
\usepackage{mathtools}
\usepackage[autostyle]{csquotes}
\usepackage{xcolor}
\usepackage{caption}
\usepackage{subcaption}
\usepackage{multirow}

\newcommand{\wrt}[1]{\mathrm{d}{#1}}
\newcommand{\KL}[2]{\mathbb{D}_{\text{KL}}\left[ #1 || #2 \right]}
\DeclareMathOperator*{\E}{\mathbb{E}}
\newcolumntype{M}[1]{>{\centering\arraybackslash}m{#1}}
\newcolumntype{N}{@{}m{0pt}@{}}
\newcolumntype{P}[1]{>{\centering\arraybackslash}p{#1}}

\colorlet{myGreen}{green!40!gray}

\aclfinalcopy 

\makeatletter
\newcommand{\thickhline}{%
    \noalign {\ifnum 0=`}\fi \hrule height 1pt
    \futurelet \reserved@a \@xhline
}
\makeatother

\title{Unsupervised Opinion Summarization  as Copycat-Review Generation}

\author{Arthur Bražinskas$^1$, Mirella Lapata$^1$ \and Ivan Titov$^{1, 2}$ \\
$^1$ ILCC, University of Edinburgh  \\
$^2$ ILLC, University of Amsterdam \\
\texttt{abrazinskas@ed.ac.uk, \{mlap,ititov\}@inf.ed.ac.uk} \\
}

\date{}

\begin{document}
\maketitle

\begin{abstract}

  Opinion summarization is the task of automatically creating
  summaries that reflect subjective information expressed in multiple
  documents, such as product reviews. While the majority of previous
  work has focused on the extractive setting, i.e.,~selecting
  fragments from input reviews to produce a summary, we let the model
  generate novel sentences and hence produce abstractive
  summaries. Recent progress in summarization has seen the development
  of supervised models which rely on large quantities of
  document-summary pairs. Since such training data is expensive to
  acquire, we instead consider the unsupervised setting, in other
  words, we do not use any summaries in training. We define a
  generative model for a review collection which capitalizes on the
  intuition that when generating a new review given a set of other
  reviews of a product, we should be able to control the ``amount of
  novelty'' going into the new review or, equivalently, vary the
  extent to which it deviates from the input. At test time, when
  generating summaries, we force the novelty to be
  minimal, and produce a text reflecting consensus opinions. We
  capture this intuition by defining a hierarchical variational
  autoencoder model. Both individual reviews and the products they
  correspond to are associated with stochastic latent codes, and the
  review generator (``decoder'') has direct access to the text of
  input reviews through the pointer-generator mechanism.  Experiments
  on Amazon and Yelp datasets, show that setting at test time the
  review's latent code to its mean, allows the model to produce
  fluent and  coherent summaries reflecting common opinions.

\end{abstract}


\section{Introduction}


\begin{table}[t!]
 	\footnotesize 
    \begin{tabular}{  >{\centering\arraybackslash} m{1.5cm}  m{5.3cm} }
     \thickhline
    \textbf{Summary} &         \vspace{0.5em}
This \textcolor{red}{restaurant is a hidden gem in Toronto}. \textcolor{orange}{The food is delicious}, and \textcolor{cyan}{the service is impeccable}. \textcolor{red}{Highly recommend} for anyone who likes \textcolor{red}{French bistro}. \vspace{0.5em} \\  \thickhline
    \textbf{Reviews} & \vspace{0.5em}\textcolor{orange}{We got the steak frites and the chicken frites both of which were very good} ... \textcolor{cyan}{Great service} ... $\vert \vert$ \textcolor{red}{I really love this place} ... \textcolor{orange}{Côte de Boeuf} ... \textcolor{red}{A Jewel in the big city}  ... $\vert \vert$ \textcolor{red}{French jewel of Spadina and Adelaide , Jules} ... \textcolor{cyan}{They are super accommodating} ... \textcolor{orange}{moules and frites are delicious} ... $\vert \vert$ \textcolor{orange}{Food came with tons of greens and fries along with my main course , thumbs uppp} ... $\vert \vert$ \textcolor{cyan}{Chef has a very cool and fun attitude} ... $\vert \vert$ \textcolor{red}{Great little French Bistro spot} ... \textcolor{red}{Go if you want French bistro food classics} ...  $\vert \vert$ \textcolor{red}{Great place} ... \textcolor{orange}{the steak frites and it was amazing} ... \textcolor{orange}{Best Steak Frites} ... \textcolor{red}{in Downtown Toronto} ... $\vert \vert$ \textcolor{red}{Favourite french spot in the city} ... \textcolor{orange}{crème brule for dessert} \vspace{0.5em}  \\\thickhline
    \end{tabular}
    \caption{A summary produced by our model; colors encode its alignment to the input reviews. The reviews are truncated, and delimited with the symbol `$\vert \vert$'. }
    \label{table:yelp_front_example_summ2}
\end{table}

Summarization of user opinions expressed in online resources, such as blogs, reviews, social media, or internet forums, has drawn much attention due to its potential for various information access applications, such as creating digests, search, and report generation~\cite{hu2004mining,angelidis2018summarizing,medhat2014sentiment}. 
Although there has been significant progress recently in summarizing
non-subjective context~\citep{rush2015neural,
  nallapati2016abstractive, paulus2017deep, see2017get,
  liu2018generating}, modern deep learning methods rely on large
amounts of annotated data that are not readily available in the
opinion-summarization domain and expensive to produce. Moreover,
annotation efforts would have to be undertaken for multiple domains as
online reviews are inherently multi-domain
~\cite{blitzer2007biographies}
and summarization systems highly
domain-sensitive~\cite{isonuma2017extractive}. Thus, perhaps
unsurprisingly, there is a long history of applying unsupervised and
weakly-supervised methods to opinion summarization
(e.g.,~\citealt{mei2007topic, titov2008modeling,
  angelidis2018summarizing}), however, these approaches have primarily
focused on extractive summarization, i.e.,~producing summaries by
copying parts of the input reviews.

In this work, we instead consider abstractive summarization which
involves generating new phrases, possibly rephrasing or using words
that were not in the original text.  Abstractive summaries are often
preferable to extractive ones as they can synthesize content across
documents avoiding redundancy
\cite{barzilay1999information,carenini2008extractive,di2014hybrid}.
In addition, we focus on the unsupervised setting and do not use any
summaries for training. Unlike aspect-based summarization
\citep{liu2012sentiment}, which rewards the diversity of opinions, we
aim to generate summaries that represent \emph{consensus}
(i.e.,~dominant opinons in reviews). We argue that such summaries can
be useful for quick decision making, and to get an overall
feel for a product or business (see the example in
Table~\ref{table:yelp_front_example_summ2}).

More specifically, we assume we are provided with a large collection
of reviews for various products and businesses and define a generative
model of this collection.  Intuitively, we want to design such a model
that, when generating a review for a product\footnote{For simplicity,
  we refer to both products (e.g., iPhone X) and businesses (e.g., a
  specific Starbucks branch) as {\it products}.} relying on a set of
other reviews, we can control the ``amount of novelty'' going into the
new review or, equivalently, vary the extent to which it deviates from
the input.  At test time, we can force the novelty to be minimal, and
generate summaries representing consensus opinions.

We capture this intuition by defining a hierarchical variational
autoencoder (VAE) model. Both products and individual reviews are
associated with latent representations.  Product representations can
store, for example, overall sentiment, common topics, and opinions
expressed about the product. In contrast, latent representations of
reviews depend on the product representations and capture the content
of individual reviews.  While at training time the latent
representations are random variables, we fix them to their respective
means at test time. As desired for summarization, these `average' (or
`copycat') reviews differ in writing style from a typical review. For
example, they do not contain irrelevant details that are common in
customer reviews, such as mentioning the occasion or saying how many
family members accompanied the reviewer. In order to encourage the
summaries to include specific details, the review generator
(`decoder') has direct access to the text of input reviews through the
pointer-generator mechanism~\cite{see2017get}.  In the example in
Table~\ref{table:yelp_front_example_summ2}, the model included
specific information about the restaurant type and its location in the
generated summary. As we will see in ablation experiments, without
this conditioning, model performance drops substantially, as the
summaries become more generic.

We evaluate our approach on two datasets, Amazon product reviews and
Yelp reviews of businesses.
%
%
The only previous method dealing with unsupervised multi-document
opinion summarization, as far as we are aware of, is
MeanSum~\cite{chu2019meansum}. Similarly to our work, they generate
consensus summaries and consider the Yelp benchmark.  Whereas we rely
on continuous latent representations, they treat the summary itself as
a discrete latent representation of a product. Although this captures
the intuition that a summary should relay key information about a
product, using discrete latent sequences makes optimization
challenging;~\cite{miao2016language,baziotis2019seq,chu2019meansum} all have to use an extra training loss term and
biased gradient estimators.

Our contributions can be summarized as follows:
\begin{itemize}
\item we introduce a simple end-to-end approach to unsupervised
  abstractive summarization;
\item we demonstrate that the approach substantially outperforms the
  previous method, both when measured with automatic metrics and in
  human evaluation;
    \item we provide a dataset of abstractive summaries for Amazon products.\footnote{Data and code: \url{https://github.com/ixlan/Copycat-abstractive-opinion-summarizer}.}
\end{itemize}


\section{Model and Estimation}

As discussed above, we approach the summarization task from a
generative modeling perspective.  We start with a high level
description of our model, then, in Sections~\ref{sec:bound}
and~\ref{sec:design}, we describe how we estimate the model and provide
extra technical details. In Section~\ref{sec:summ_gen}, we explain how
we use the model to generate summaries.

\subsection{Overview of the Generative Model}

\begin{figure*}[t!]
\centering
\begin{subfigure}[t]{.45\linewidth}
  \centering
  \includegraphics[width=1.\linewidth]{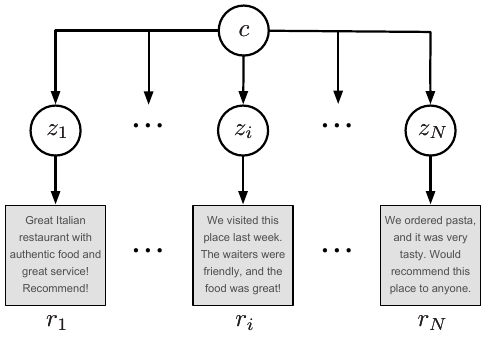}
  \caption{Conditional independence of the reviews given the group representation $c$.}
  \label{fig:unfolded_gm_base}
\end{subfigure}%
\quad
\begin{subfigure}[t]{.45\linewidth}
  \centering
  \includegraphics[width=1.\linewidth]{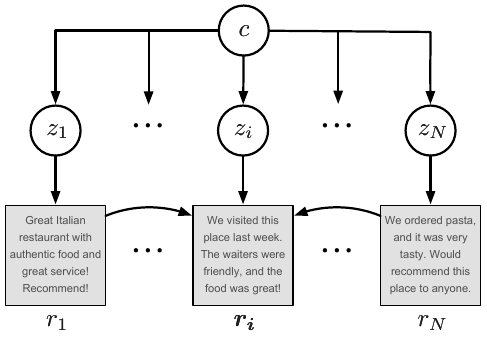}
  \caption{The $r_i$'s decoder accesses other reviews of the group ($r_1$, ..., $r_{i-1}$, $r_{i+1}$, ..., $r_N$).}
  \label{fig:unfolded_gm_copy}
\end{subfigure}
\caption{Unfolded graphical representation of the model.}
\label{fig:unfolded_gm}
\end{figure*}

Our text collection consists of groups of reviews, with each group
corresponding to a single product. Our latent
summarization model (which we call \textsc{CopyCat}) captures
this hierarchical organization and can be regarded as an extension of
the vanilla text-VAE model~\cite{bowman2015generating}.
\textsc{CopyCat} uses two sets of latent variables as shown in
Figure~\ref{fig:unfolded_gm_base}. Namely, we associate each review
group (equivalently, each product) with a continuous variable~$c$,
which captures the group's `latent semantics'.  In addition, we
associate each individual review ($r_i$) with a continuous variable
$z_i$, encoding the semantics of that review.
The information stored in $z_i$ is used by the
decoder~$p_{\theta}(r_i| z_i)$ to produce review text~$r_i$. The
marginal log-likelihood of one group of reviews~$r_{1:N} = (r_1, \ldots, r_N)$ is given by
\begin{equation*}
\begin{aligned}
    &\log p_{\theta}(r_{1:N}) = \\
    &\log \int \left[ p_{\theta}(c) \prod_{i=1}^N  \left[ \int p_{\theta}(r_i| z_i) p_{\theta}(z_i|c) \wrt z_i \right] \wrt c \right],
\end{aligned}
    \label{eq:ll}
\end{equation*}
where we marginalize over variables $c$ and $z_{1:N}$.


When generating a new review $r_i$, given the set of previous reviews  $r_{1:i}$,
the information about these reviews has to be conveyed through the latent  representations $c$ and $z_i$.
This bottleneck is undesirable, as it will make it hard
for the model to pass fine-grain information. For example, at generation time, the model should 
be reusing
 named entities (e.g., product names or technical characteristics)  from other reviews rather than  `hallucinating' or avoiding generating them at all, resulting in  
 generic and non-informative text.
%
%
%
We alleviate this issue
by letting the decoder directly access other reviews.
We can formulate this as an autoregressive model:
\begin{equation}
\begin{aligned}
p_{\theta}(r_{1:N} | c) = \prod_{i=1}^N p_{\theta}(r_i| r_1,...,r_{i-1}, c).
\end{aligned}
    \label{eq:chain_rule}
\end{equation}
As we discuss in Section~\ref{sec:design}, the conditioning is
instantiated using the pointer-generator mechanism~\cite{see2017get}
and, thus, will specifically help in generating rare words
(e.g.,~named entities).

We want our summarizer to equally rely on every review, without
imposing any order (e.g.,~temporal) on the generation process.
Instead, as shown in Figure~\ref{fig:unfolded_gm_copy}, when
generating $r_i$, we let the decoder access all other reviews within a
group, $r_{-i} = (r_1, \ldots, r_{i-1}, r_{i+1}, \ldots, r_N)$. This
is closely related to pseudolikelihood
estimation~\cite{besag1975statistical} or Skip-Thought's
objective~\cite{kiros2015skip}.
The final objective that we maximize for each group of reviews $r_{1:N}$:
\begin{equation}
\hspace*{-.26cm}\log\hspace*{-.2ex}\!\hspace{-.2ex}\int\hspace{-.4ex}p_{\theta}(c)
\hspace*{-.5ex}\prod_{i=1}^N  \left[ \int\hspace*{-1ex}p_{\theta}(r_i|
  z_i, r_{\_i})   \right.  \left.  p_{\theta}(z_i|c) \wrt z_i
  \vphantom{\int} \right] \hspace*{-.5ex}\wrt c \vphantom{\prod_{i=1}^N}\;
\label{eq:actual_ll}
\end{equation}
We will confirm in ablation experiments that both hierarchical
modeling (i.e.,~using $c$) and the direct conditioning on other
reviews are beneficial.

\subsection{Model Estimation}
\label{sec:bound}

As standard with VAEs and variational inference in general~\cite{kingma2013auto},
instead of directly maximizing  the intractable marginal likelihood
 in Equation~\ref{eq:actual_ll}, we maximize its lower bound:\footnote{See the derivations in Appendix \ref{sec:elbo_der}.}
\begin{equation*}
\begin{aligned}
  &\mathcal{L}(\theta, \phi; r_{1:N}) = \\
  &\E_{c \sim q_{\phi}(c | r_{1:N})} \! \! \left[ \sum_{i=1}^N \E_{z_i \sim q_\phi(z_i | r_i, c)} \!
 \! \left[ \log p_{\theta}(r_i | z_i, r_{\_i}) \right] \right.\\
   &- \left. \vphantom{} \sum_{i=1}^N  \KL{q_{\phi}(z_i| r_i, c)}{p_{\theta}(z_i| c)} \right] \\
    &- \vphantom{} \KL{q_{\phi}(c | r_{1:N})}{p_{\theta}(c)}.
\end{aligned}
\end{equation*}
The lower bound includes two `inference networks',
$q_{\phi}(c | r_{1:N})$ and $q_\phi(z_i | r_i, c)$, which are neural
networks parameterized with $\phi$ and will be discussed in detail in
Section~\ref{sec:design}.  They approximate the corresponding
posterior distributions of the model.  The first term is the
reconstruction error: it encourages the quality reconstruction of the
reviews.  The other two terms are regularizers. They control the
amount of information encoded in the latent representation by
penalizing the deviation of the estimated posteriors from the
corresponding priors, the deviation is measured in terms of the
Kullback-Leibler (KL) divergence.  The bound is maximized with respect
to both the generative model's parameters~$\theta$ and inference
networks' parameters~$\phi$.  Due to Gaussian assumptions, the
Kullback-Leibler (KL) divergence terms are available in closed form,
while we rely on the reparameterization trick~\cite{kingma2013auto} to
compute gradients of the reconstruction term.

The inference network predicting the posterior 
for a review-specific variable $q_\phi(z_i | r_i, c)$ is needed only in training and  is discarded afterwards. In contrast, we will exploit the  
inference network $q_{\phi}(c | r_{1:N})$
when generating summaries, as discussed in Section~\ref{sec:summ_gen}.

\subsection{Design of Model Components}
\label{sec:design}


\subsubsection{Text Representations}
A GRU encoder \citep{cho2014learning} embeds review words $w$ to
obtain hidden states $h$. Those representations are reused across the
system, e.g., in the inference networks and the decoder.


\vspace{1ex}

\begin{figure}[t]
    \centering
    \includegraphics[width=0.48\textwidth]{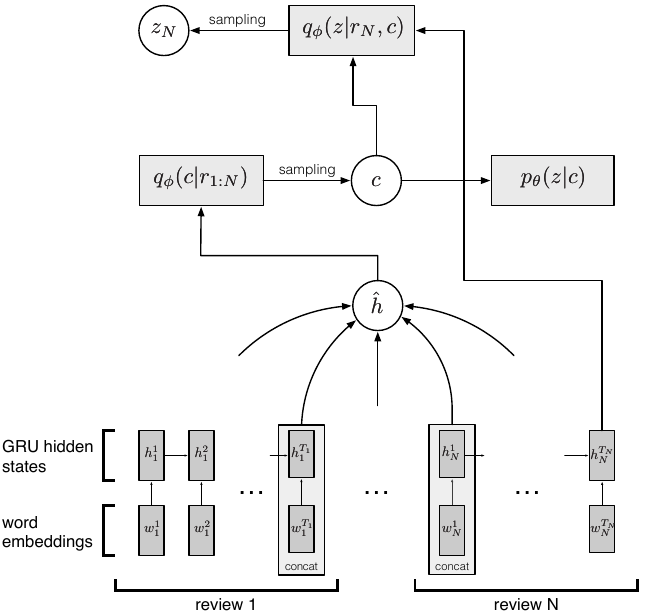}
    \caption{Production of  latent code~$z_N$ for review~$r_N$.}
    \label{fig:latent_arch}
\end{figure}

\noindent 
The full architecture used to produce the latent codes $c$ and $z_i$ is shown in Figure \ref{fig:latent_arch}. We make Gaussian assumptions for all distributions (i.e. posteriors and priors).  As in \citet{kingma2013auto}, we use separate linear projections (LPs) to compute the means and diagonal log-covariances.


\subsubsection{Prior $p(c)$ and posterior  $q_{\phi}(c | r_{1:N})$}
We set the prior over group latent codes to the standard normal distribution, $p(c)=\mathcal{N}(c; 0, I)$.
In order to compute the approximate posterior $q_{\phi}(c | r_{1:N})$,
we first predict the contribution (`importance') of each word in each review $\alpha_i^t$ to the code of the group:
\begin{equation*}
    \alpha_i^t = \dfrac{\exp(f_{\phi}^\alpha(m_i^t))}
                {\sum_{j=1}^N \sum_k^{T_j} \exp(f_{\phi}^\alpha(m_j^k))},
    \label{eq:alpha}
\end{equation*} where 
 $T_i$ is the length of $r_i$ and $f_{\phi}^{\alpha}$ is a feed-forward neural network (FFNN)\footnote{We use FFNNs with the $\tanh$ non-linearity in several model components. Whenever a FFNN is mentioned in the subsequent discussion, this architecture is assumed.} which takes
 as input concatenated word embeddings and hidden states of the GRU encoder, $m_{i}^t = [h_{i}^t \circ w_{i}^t]$, and returns a scalar.


 
Next, we compute the intermediate representation with the weighted sum: $  \hat h = \sum_{i=1}^N \sum_t^{T_i}  \alpha_i^t m_i^t$. 
Finally, we compute the Gaussian's parameters using the affine projections:
\begin{align*}
    \mu_{\phi}(r_{1:N}) = L\hat h + b_L \\
    \log \sigma_{\phi}(r_{1:N}) = G\hat h + b_G
\end{align*}

\subsubsection{Prior $p_{\theta}(\!z_i | c\!)$ and posterior $q_{\phi}(z_i |r_i,\! c)$}
To compute the prior on the review code $z_i$, $p_{\theta}(z_i | c) = \mathcal{N}(z_i; \mu_{\theta}(c), I\sigma_{\theta}(c))$, we linearly project the product code $c$. Similarly, to compute the parameters of the approximate posterior $q_{\phi}(z |r_i, c)$ = $\mathcal{N}(z; \mu_{\phi}(r_i, c), I\sigma_{\phi}(r_i, c))$, we concatenate the last encoder's state $h_i^{T_i}$ of the review $r_i$ and $c$, and perform  affine transformations. 

\subsubsection{Decoder $p_{\theta}(r_i | z_i, r_{\_i})$}
To compute the distribution $p_{\theta}(r_i | z_i, r_{\_i})$, we use an auto-regressive GRU decoder with the attention mechanism \citep{bahdanau2014neural} and a pointer-generator network.

We compute the context vector $c_i^t = \text{att}(s_i^t, h_{\_i})$ by attending to all the encoder's hidden states $h_{\_i}$ of the other reviews $r_{\_i}$ of the group, where the decoder's hidden state $s_i^t$ is used as a query. The hidden state of the decoder is computed using the GRU cell as 
\begin{equation}
    s_i^t = \text{GRU}_{\theta}(s_i^{t-1}, [ w_i^t \circ c_i^{t-1} \circ z_i ]).
\end{equation}

The cell inputs the previous hidden state $s_i^{t-1}$, as well as concatenated word embedding $w_i^t$, context vector $c_i^{t-1}$, and  latent code $z_i$. 

Finally, we compute the word distributions using the pointer-generator network $g_{\theta}$:
\begin{equation}
    p_{\theta}(r_i | z_i, r_{\_i}) = \prod_{t=1}^T g_{\theta}(r_i^t | s_i^t, c_i^t, w_i^t, r_{\_i})
\end{equation}

The pointer-generator network computes two internal word distributions that are hierarchically aggregated  into one distribution \citep{morin2005hierarchical}. One distribution assigns probabilities to words being generated using a fixed vocabulary, and another one probabilities to be copied directly from the other reviews $r_{\_i}$. In our case, the network helps to preserve details and, especially, to generate rare tokens.

\section{Summary Generation}
\label{sec:summ_gen}
Given reviews~$r_{1:N}$, we generate a summary that reflects common
information using trained components of the model. Formally, we could
sample a new review from
\begin{equation}
\begin{aligned}
\nonumber
&p_{\theta}(r| r_{1:N}) = \\
&\E_{c \sim q_{\phi}(c| r_{1:N})} \left[ \E_{z \sim p_{\theta}(z|c)} \left[ p_{\theta}(r| z, r_{1:N}) \right] \right].
\end{aligned}
\end{equation}

As we argued in the introduction and will revisit in experiments, a
summary or summarizing review, should be generated relying on the mean
of the reviews' latent code. Consequently, instead of sampling $z$ from
$p_{\theta}(z | c) = \mathcal{N}(z; \mu_{\theta}(c),
I\sigma_{\theta}(c))$,
we set it to $\mu_{\theta}(c)$. 
We also found beneficial, in terms of evaluation metrics, not to sample $c$ but instead to rely on the mean
predicted by the inference network $q_{\phi}(c| r_{1:N})$.

\section{Experimental Setup}
\subsection{Datasets}
Our experiments were conducted on business customer reviews from the
Yelp Dataset Challenge and Amazon product reviews
\citep{he2016ups}. These were pre-processed similarly to
\citet{chu2019meansum}, and the corresponding data statistics are
shown in Table~\ref{table:data_stats}. Details of the
pre-processing are available in  Appendix~\ref{sec:dataset_prep}.

These datasets present different challenges to abstractive
summarization systems. Yelp reviews contain much personal information
and irrelevant details which one may find unnecessary in a
summary. Our summarizer, therefore, needs to distill important
information in reviews while abstracting away from details such as
a listing of all items on the menu, or mentions of specific dates or
occasions upon which customers visited a restaurant.
On the contrary, in Amazon reviews, we observed that users tend to
provide more objective information and specific details that are
useful for decision making (e.g.,~the version of an electronic
product, its battery life, its dimensions). In this case, it would be
desirable for our summarizer to preserve this information in the
output summary.

\begin{table}
\centering
    \begin{tabular}{ c  c  c  c }     \thickhline
    Dataset & Training & Validation \\ \thickhline
    Yelp & 38,776/1,012,280 & 4,311/113,373\\ 
    Amazon & 183,103/4,566,519 & 9,639/240,819\\ \thickhline
    \end{tabular}
    \caption{Data statistics after pre-processing. The format in the cells is Businesses/Reviews and Products/Reviews for Yelp and Amazon, respectively.}
    \label{table:data_stats}
\end{table}

For evaluation, we used the same 100 human-created Yelp summaries
released by \citet{chu2019meansum}. These were generated by Amazon
Mechanical Turk (AMT) workers, who summarized 8 input reviews. We created a new
test for Amazon reviews following a similar procedure (see
Appendix~\ref{sec:ama_summs_creation} for details). We sampled 60
products and 8 reviews for each product, and they were shown to AMT workers who
were asked to write a summary. We collected three summaries per
product, 28~products were used for development and 32~for testing.


\subsection{Experimental Details}
\label{sec:exp_det}
We used GRUs \citep{cho2014learning} for sequential encoding and
decoding we used GRUs. We randomly initialized word embeddings that
were shared across the model as a form of regularization
\citep{press2016using}. Further, optimization was performed using Adam
\citep{kingma2014adam}. 
In order to overcome the \enquote{posterior collapse}
\citep{bowman2015generating}, 
both for our model and the vanilla VAE baseline, 
we applied cyclical annealing \citep{liu2019cyclical}. The reported
ROUGE scores are based on F1 (see Appendix~\ref{sec:hparams} for
details on hyperparameters).

\subsection{Baseline Models}
\textit{Opinosis} is a graph-based abstractive summarizer
\citep{ganesan2010opinosis} designed to generate short opinions based
on highly redundant texts. Although it is referred to as abstractive,
it can only select words from the reviews.

\textit{LexRank} is an unsupervised algorithm which selects sentences
to appear in the summary based on graph centrality (sentences
represent nodes in a graph whose edges have weights denoting
similarity computed with tf-idf).  A node's centrality can be
measured by running a ranking algorithm such as PageRank
\citep{page1999pagerank}.

\textit{MeanSum}\footnote{For experiments on Yelp, we used the
  checkpoint provided by the authors, as we obtained very similar
  ROUGE scores when retraining the model.}  is the unsupervised
abstractive summarization model~\cite{chu2019meansum} discussed in the
introduction.

We also trained a vanilla text \textit{VAE} model
\citep{bowman2015generating} with our GRU encoder and decoder. When
generating a summary for $r_1, ..., r_N$, we averaged the means of
$q_{\phi}(z_i|r_i)$.

Finally, we used a number of simple summarization baselines. We
computed the \textit{clustroid} review for each group as follows. We
took each review from a group and computed ROUGE-L with respect to all
other reviews.  The review with the highest ROUGE score was selected
as the clustroid review.  Furthermore, we sampled a \textit{random}
review from each group as the summary, and constructed the summary by
selecting the \textit{leading sentences} from each review of a group.

Additionally, as an upper bound, we report the performance of an
\textit{oracle} review, i.e., the highest-scoring review in a group
when computing  ROUGE-L against reference summaries. 

\begin{table}[t]
\centering
    \begin{tabular}{  l  c  c  c }\thickhline
     & R1 & R2 & RL \\ \thickhline
    Copycat & \textbf{0.2947} & \textbf{0.0526} & \textbf{0.1809} \\ 
    MeanSum & 0.2846 & 0.0366  & 0.1557 \\ 
    LexRank & 0.2501 & 0.0362 & 0.1467 \\ 
    Opinosis & 0.2488 & 0.0278 & 0.1409 \\ 
    VAE & 0.2542 & 0.0311  & 0.1504 \\ \thickhline
    Clustroid & 0.2628 & 0.0348 & 0.1536 \\ 
    Lead & 0.2634 & 0.0372 & 0.1386 \\ 
    Random & 0.2304 & 0.0244 & 0.1344 \\ \thickhline
    Oracle & 0.2907 & 0.0527 & 0.1863 \\ \thickhline 
    \end{tabular}
    \caption{ROUGE scores on the Yelp test set.}
    \label{table:yelp_rouge}
\end{table}


\section{Evaluation Results}
\subsection{Automatic Evaluation}
As can be seen in Tables~\ref{table:yelp_rouge}
and~\ref{table:amazon_rouge}, our model, Copycat, yields the highest
scores on both Yelp and Amazon datasets.

We observe large gains over the vanila VAE.  We conjecture
that the vanilla VAE struggles to properly represent the variety of
categories under a single prior $p(z)$. For example, reviews about a
sweater can result in a summary about socks (see example summmaries in
Appendix).
This contrasts with our model which allows each group to have its own
prior 
$p_{\theta}(z|c)$ and access to other reviews during
decoding. 
The gains are especially large on the Amazon dataset, which is very
broad in terms of product categories.

Our model also substantially outperforms MeanSum. As we will confirm
in human evaluation, MeanSum's summaries are relatively fluent at the
sentence level but often contain hallucinations, i.e.,~information not
present in the input reviews.


\begin{table}[t!]
\centering
    \begin{tabular}{  l  c  c  c }     \thickhline
     & R1 & R2 & RL \\ \thickhline
    Copycat & \textbf{0.3197} & \textbf{0.0581} & \textbf{0.2016} \\ 
    MeanSum & 0.2920 & 0.0470 & 0.1815 \\ 
    LexRank & 0.2874 & 0.0547 & 0.1675 \\ 
    Opinosis & 0.2842 & 0.0457 & 0.1550\\ 
    VAE & 0.2287 & 0.0275 & 0.1446 \\ \thickhline
    Clustroid  & 0.2928 & 0.0441 & 0.1778 \\ 
    Lead & 0.3032 & \textbf{0.0590} & 0.1578 \\ 
    Random & 0.2766 & 0.0472 & 0.1695 \\ \thickhline
    Oracle & 0.3398 & 0.0788 & 0.2160 \\ \thickhline 
    \end{tabular}
    \caption{ROUGE scores on the Amazon test set.}
    \label{table:amazon_rouge}
\end{table}

\begin{table*}[t]
\centering
\begin{tabular}{ l c  c  c  c  c }
    \hline
       & Fluency & Coherence & Non Red. & Opinion Cons. & Overall \\ \thickhline
       Copycat & \textbf{0.5802} & \textbf{0.5161} & \textbf{0.4722} & -0.0909 & \textbf{0.3818} \\ 
       MeanSum & -0.5294 & -0.4857 & 0.0270 & -0.6235 & -0.7468 \\
       LexRank & -0.7662 & -0.8293 & -0.7699 & \textbf{0.3500} & -0.5278 \\ \thickhline
         Gold & 0.6486 & 0.8140 & 0.6667 & 0.3750 & 0.8085 \\ \thickhline
\end{tabular}
    \caption{Human evaluation results in terms of the Best-Worst scaling on the Yelp dataset.}
    \label{table:yelp_human_eval}
\end{table*}

\begin{table*}[t]
\centering
\begin{tabular}{ l  c  c  c  c  c } \thickhline
       & Fluency & Coherence & Non Red. & Opinion Cons. & Overall \\ \thickhline
       Copycat & \textbf{0.4444} & \textbf{0.3750} & \textbf{0.0270} & -0.4286 & -0.1429\\ 
       MeanSum & -0.6410 & -0.8667 & -0.6923 & -0.7736 & -0.8305\\ 
       LexRank & -0.2963 & -0.3208 & -0.3962 & \textbf{0.4348} & \textbf{0.1064} \\ 
       \thickhline
       Gold & 0.3968 & 0.7097 & 0.7460 & 0.6207 & 0.7231 \\ \thickhline
\end{tabular}
    \caption{Human evaluation results in terms of the Best-Worst scaling on the Amazon dataset.}
    \label{table:ama_human_eval}
\end{table*}

\subsection{Human Evaluation}
\paragraph{Best-Worst Scaling}
\label{sec:bw}
We performed human evaluation using the AMT platform. We
sampled 50 businesses from the human-annotated Yelp test set and used
all 32 test products from the Amazon set. We recruited 3 workers to
evaluate each tuple containing summaries from MeanSum, our model,
LexRank, and human annotators. The reviews and summaries were
presented to the workers in random order and were judged using
Best-Worst Scaling \citep{louviere1991best, louviere2015best}. BWS has
been shown to produce more reliable results than ranking scales
\citep{kiritchenko2017capturing}. Crowdworkers were asked to judge
summaries according to the criteria listed below (we show an abridged
version below, the full set of instructions is given in Appendix
\ref{sec:full_he_instr}). The non-redundancy and coherence criteria
were taken from \citet{dang2005overview}.

    \textit{Fluency}: the summary sentences should be grammatically correct, easy to read and understand; 
    \textit{Coherence}: the summary should be well structured and well organized;
     \textit{Non-redundancy}: there should be no unnecessary repetition in the summary;
     \textit{Opinion consensus}: the summary should reflect common opinions expressed in the reviews; 
    \textit{Overall}: based on your own criteria (judgment) please select the best and the worst summary of the reviews.

For every criterion, a system’s score is computed as the percentage of times it was selected as best minus the percentage of times it was selected as worst \citep{orme2009maxdiff}. The scores range from -1 (unanimously worst) to +1 (unanimously best).

On Yelp, as shown in Table \ref{table:yelp_human_eval}, our model
scores higher than the other models according to most criteria, including
overall quality.
The differences with other systems are statistically significant for
all the criteria at $p < 0.01$, using post-hoc HD Tukey tests. The
difference in fluency between our system and gold summaries is not
statistically significant.

The results on Amazon are shown in
Table~\ref{table:ama_human_eval}. Our system outperforms other methods
in terms of fluency, coherence, and non-redundancy. As with Yelp, it
trails LexRank according to the opinion consensus
criterion. Additionally, LexRank is slightly preferable overall.
All pairwise differences between our model and comparison systems are
statistically significant at $p < 0.05$.



\textit{Opinion consensus} (OC) is a criterion that captures the
coverage of common opinions, and it seems to play a different role in
the two datasets. On Yelp, LexRank has better coverage compared to our
model, as indicated by the higher OC score, but is not preferred
\textit{overall}. In contrast, on Amazon, while the OC score is on the
same par, LexRank is preferred \textit{overall}.  We suspect that
presenting a breadth of exact details on Amazon is more important than on
Yelp. 
Moreover, LexRank tends to produce summaries that are about 20
tokens longer than ours resulting in better coverage of input details.

\paragraph{Content Support}
\label{sec:cs}
The ROUGE metric relies on unweighted n-gram overlap and can be
insensitive to hallucinating facts and entities \citep{falke2019ranking}. For example,
referring to a burger joint as a veggie restaurant is highly
problematic from a user perspective but yields only marginal
differences in ROUGE.  To investigate how well the content of the
summaries is supported by the input reviews, we performed a second
study. We used the same sets as in the human evaluation in
Section~\ref{sec:bw}, and split MeanSum and our system's summaries
into sentences. Then, for each summary sentence, we assigned 3
AMT workers to assess how well the sentence is supported
by the reviews.  Workers were advised to read the reviews and rate
sentences using one of the following three options. \textit{Full
  support}: all the content is reflected in the reviews;
\textit{Partial support}: only some content is reflected in the
reviews; \textit{No support}: content is not reflected in the reviews.


%

The results in Table~\ref{table:content_support} indicate that our
model is better at preserving information than MeanSum.


\section{Analysis}
\paragraph{Ablations}

To investigate the importance of the model's individual components, we
performed ablations by removing the latent variables ($z_i$ and $c$,
one at a time), and attention over the other
reviews. The models were re-trained on the Amazon dataset. The results
are shown in Table~\ref{table:amazon_ablation_rouge}. They indicate
that all components play a role, yet the most significant drop in
ROUGE was achieved when the variable $z$ was removed, and only $c$
remained. Summaries obtained from the latter system were wordier and
looked more similar to
reviews. 
Dropping the attention (w/o $r_{\_i}$) results in more generic
summaries as the model cannot copy details from the input. Finally,
the smallest quality drop in terms of ROUGE-L was observed when
the variable $c$ was removed.

In the introduction, we hypothesized that using the mean of latent
variables would result in more ``grounded'' summaries reflecting the
content of the input reviews, whereas sampling would yield texts with
many novel and potentially irrelevant details. To empirically
test this hypothesis, we sampled the latent variables during summary
generation, as opposed to using mean values (see Section
\ref{sec:summ_gen}). We indeed observed that the summaries were
wordier, less fluent, and less aligned to the input reviews, as is
also reflected in the ROUGE scores
(Table~\ref{table:amazon_ablation_rouge}).

\begin{table}[t]
\begin{tabular}{@{}l@{~}c@{\hspace{1ex}}c@{\hspace{1.5ex}}c@{\hspace{1ex}}c@{}} \thickhline
     &        \multicolumn{2}{c}{Yelp} & \multicolumn{2}{c}{Amazon}\\\thickhline
     &         Copycat & MeanSum  & Copycat & MeanSum\\
Full & \textbf{44.50} & 28.41 & \textbf{38.23} & 24.41  \\
Partial &   \textbf{32.48} & 30.66 &\textbf{33.95} & 31.23  \\
No &\textbf{23.01}&  40.92 & \textbf{27.83} &  44.36 \\ \thickhline
\end{tabular}
\caption{Content support on Yelp and Amazon datasets, percentages.}
\label{table:content_support}
\end{table}

\paragraph{Copy Mechanism}
Finally, we analyzed which words are copied by the full model during summary
generation. 
Generally, the model copies around 3-4 tokens per
summary. 
We observed a tendency to copy product-type specific words (e.g.,
\textit{shoes}) as well as brands and names.


\section{Related Work}
\label{sec:related_work}

Extractive weakly-supervised opinion summarization has been an active
area of research.  A recent example is
\citet{angelidis2018summarizing}. First, they learn to assign
sentiment polarity to review segments in a weakly-supervised
fashion. Then, they induce aspect labels for segments relying on a
small sample of gold summaries. Finally, they use a heuristic to
construct a summary of segments.  Opinosis \citep{ganesan2010opinosis}
does not use any supervision. The model relies on redundancies in
opinionated text and PoS tags in order to generate short
opinions. This approach is not well suited for the generation of
coherent long summaries and although it can recombine fragments of
input text, it cannot generate novel words and phrases.  LexRank
\citep{erkan2004lexrank} is an unsupervised extractive approach which
builds a graph in order to determine the importance of sentences, and
then selects the most representative ones as a summary. 
\citet{isonuma2019unsupervised} introduce an unsupervised approach for
single review summarization, where they rely on latent discourse
trees.  Other earlier approaches \citep{gerani2014abstractive,
  di2014hybrid} relied on text planners and templates, while our
approach does not require rules and can produce fluent and varied
text.  Finally, conceptually related methods were applied to
unsupervised single sentence compression \citep{west2019bottlesum,
  baziotis2019seq, miao2016language}.  The most related approach to
ours is MeanSum~\citep{chu2019meansum} which treats a summary as a
discrete latent state of an autoencoder. In contrast, we define a
hierarchical model of a review collection and use continuous latent
codes.

\begin{table}[t]
\centering
    \begin{tabular}{  l  c  c c }
    \thickhline
     & R1 & R2 & RL \\ \thickhline
    w/o $r_{\_i}$ & 0.2866 & 0.0454 &  0.1863\\ 
    w/o $c$ & \textbf{0.2767} & 0.0507 & 0.1919  \\ 
    w/o $z$ & 0.2926 &\textbf{0.0416} & \textbf{0.1739} \\ 
    Sampling & 0.2563 & 0.0434 & 0.1716 \\ \thickhline
    Full & 0.3197 & 0.0581 & 0.2016 \\ \thickhline
    \end{tabular}
    \caption{Ablations, ROUGE scores on Amazon.} 
    \label{table:amazon_ablation_rouge}
\end{table}




\section{Conclusions}
In this work, we presented an abstractive summarizer of opinions,
which does not use any summaries in training and is trained end-to-end
on a large collection of reviews.  The model compares favorably to the
competitors, especially to the only other unsupervised abstractive
multi-review summarization system. Furthermore, human evaluation of
the generated summaries (by considering their alignment with the
reviews) shows that those created by our model better reflect the
content of the input.


\section*{Acknowledgments}
We would like to thank the anonymous reviewers
for their valuable comments. Also, Stefanos Angelidis for help with the data as well as Jonathan Mallinson, Serhii Havrylov, and other members of Edinburgh NLP group for discussion.
We gratefully acknowledge the support of the European Research
Council (Titov: ERC StG BroadSem 678254; Lapata: ERC CoG TransModal 681760) and the Dutch National Science Foundation (NWO VIDI 639.022.518).

\bibliography{bibl}
\bibliographystyle{acl_natbib}
\newpage
\appendix

\section{Appendices}
\label{sec:appendix}

\subsection{Derivation of the Lower Bound}
\label{sec:elbo_der}

To make the notation below less cluttered, we make a couple of simplifications: $q_{\phi}(c | \cdot) = q_{\phi}(c | r_{1:N})$ and $q_{\phi}(z | i) = q_{\phi}(z | r_i, c)$.
\begin{equation}
\begin{aligned}
&\log \int \left[ p_{\theta}(c) \prod_{i=1}^N p_{\theta}(r_i| c, r_{\_i}) \wrt c \right] \\
&= \log \int \left[ p_{\theta}(c) \prod_{i=1}^N  \left[ \int p_{\theta}(r_i, z| c, r_{\_i}) \wrt z \right] \wrt c \right] \\ &= 
\log \int \left[ p_{\theta}(c) \dfrac{q_{\phi}(c |\cdot)}{q_{\phi}(c| \cdot)} \prod_{i=1}^N  \left[ \int p_{\theta}(r_i, z| c, r_{\_i}) \right. \right. \\
&\left. \left. \dfrac{q_{\phi}(z | i)}{q_{\phi}(z | i)} \wrt z \right] \wrt c \right] = \log \\&\E_{c \sim q_{\phi}(c | \cdot)} \left[ \dfrac{p_{\theta}(c)}{q_{\phi}(c |\cdot)} \prod_{i=1}^N\E_{z \sim q_\phi(z | i)} \left[ \dfrac{p_{\theta}(r_i, z| c, r_{\_i})}{q_{\phi}(z | i)} \right] \right] \\
& \geq  \E_{c \sim q_{\phi}(c | \cdot)} \left[ \sum_{i=1}^N \log \E_{z \sim q_\phi(z | i)} \left[ \dfrac{p_{\theta}(r_i, z| c, r_{\_i})}{q_{\phi}(z | i)} \right] \right] \\ 
&- \KL{q_{\phi}(c | \cdot))}{p_{\theta}(c)} \geq \\
& \E_{c \sim q_{\phi}(c |\cdot)} \left[ \sum_{i=1}^N \E_{z \sim q_\phi(z | i)} \left[ \log \dfrac{p_{\theta}(r_i, z| c, r_{\_i})}{q_{\phi}(z | i)} \right] \right] - \\
&\vphantom{} \KL{q_{\phi}(c | \cdot))}{p_{\theta}(c)} = \\ 
&\E_{c \sim q_{\phi}(c | \cdot)} \left[ \sum_{i=1}^N \E_{z \sim q_\phi(z | i)} \left[ \log p_{\theta}(r_i | z, r_{\_i})  \right] \right. - \\ 
&\left. \sum_{i=1}^N  \KL{q_{\phi}(z| i)}{p_{\theta}(z| c)} \right] - \\ &\KL{q_{\phi}(c | \cdot)}{p_{\theta}(c)}
\end{aligned}
\label{eq:hvae_eblo1}
\end{equation}

 \subsection{Dataset Pre-Processing}
 \label{sec:dataset_prep}
 We selected only businesses and products with a minimum of 10 reviews, and thee minimum and maximum length of 20 and 70 words respectively, popular groups above the $90^{th}$ percentile were removed. And each group was set to contain 8 reviews during training. From the Amazon dataset we selected 4 categories: \textit{Electronics}; \textit{Clothing, Shoes and Jewelry}, \textit{Home and Kitchen}; \textit{Health and Personal Care}. 

\subsection{Hyperparameters}
\label{sec:hparams}

For sequential encoding and decoding, we used GRUs \citep{cho2014learning} with 600-dimensional hidden states. The word embeddings dimension was set to 200, and they were shared across the model \citep{press2016using}. The vocabulary size was set to 50,000 most frequent words, and an extra 30,000 were allowed in the extended vocabulary, the words were lower-cased. We used the Moses' \citep{koehn2007moses} reversible tokenizer and truecaser. Xavier uniform initialization \citep{glorot2010understanding} of 2D weights was used, and 1D weights were initialized with the scaled normal noise ($\sigma=0.1$). We used the Adam optimizer \citep{kingma2014adam}, and set the learning rate to 0.0008 and 0.0001 on Yelp and Amazon, respectively. For summary decoding, we used length-normalized beam search of size 5, and relied on latent code means. In order to overcome \enquote{posterior collapse} \citep{bowman2015generating} we applied cycling annealing \citep{liu2019cyclical} with $r=0.8$ for both the $z$ and $c$ related KL terms, with a new cycle over approximately every 2 epochs over the training set. The maximum annealing scalar was set to 1 for $z$-related KL term in on both datasets, and 0.3 and 0.65 for $c$-related KL-term on Yelp and Amazon, respectively. The reported ROUGE scores are based on F1. 

The dimensions of the variables $c$ and $z$ were set to 600, and the $c$ posterior's scoring neural network had a 300-dimensional hidden layer and the tanh non-linearity.  

The decoder's attention mechanism used a single layer neural network with a 200-dimensional hidden layer, and the \textit{tanh} non-linearity. The copy gate in the pointer-generator network was computed with a 100-dimensional single-hidden layer network, with the same non-linearity. 

\subsection{Human Evaluation Setup}
\label{sec:he_setup}

To perform the human evaluation experiments described in Sections \ref{sec:bw} and \ref{sec:cs} we combined both tasks into single Human Intelligence Tasks (HITs). Namely, the workers needed to mark sentences as described in Section \ref{sec:cs}, and then proceed to the task in Section \ref{sec:bw}. We explicitly asked then to re-read the reviews before each task. 

For worker requirements we set 98\% approval rate, 1000+ HITS, Location: USA, UK, Canada, and the maximum score on a qualification test that we designed. The test was asking if the workers are native English speakers, and verifying that they correctly understand the instructions of both tasks by completing a mini version of the actual HIT.

\subsection{Full Human Evaluation Instructions}
\label{sec:full_he_instr}

\begin{itemize}
    \item \textbf{Fluency}: The summary sentences should be grammatically correct, easy to read and understand. 
    \item \textbf{Coherence}: The summary should be well structured and well organized. The summary should not just be a heap of related information, but should build from sentence to sentence to a coherent body of information about a topic. 
     \item \textbf{Non-redundancy}:  There should be no unnecessary repetition in the summary. Unnecessary repetition might take the form of whole sentences that are repeated, or repeated facts, or the repeated use of a noun or noun phrase (e.g., "Bill Clinton") when a pronoun ("he") would suffice. 
     
     \item \textbf{Opinion consensus}: The summary should reflect common opinions expressed in the reviews. For example, if many reviewers complain about a musty smell in the hotel's rooms, the summary should include this information. 
    \item \textbf{Overall}: Based on your own criteria (judgment) please select the best and the worst summary of the reviews. 
\end{itemize}

\subsection{Amazon Summaries Creation}
\label{sec:ama_summs_creation}

First, we sampled 15 products from each of the Amazon review categories: \textit{Electronics}; \textit{Clothing, Shoes and Jewelry}; \textit{Home and Kitchen}; \textit{Health and Personal Care}. Then, we selected 8 reviews from each product to be summaries. We used the same requirements for workers as for human evaluation in \ref{sec:he_setup}. We assigned 3 workers to each product, and instructed them to read the reviews and produce a summary text. We followed the instructions provided in \citep{chu2019meansum}, and used the following points in our instructions:

\begin{itemize}
    \item The summary should reflect common opinions about the product expressed in the reviews. Try to preserve the common sentiment of the opinions and their details (e.g. what exactly the users like or dislike). For example, if most reviews are negative about the sound quality, then also write negatively about it. Please make the summary coherent and fluent in terms of sentence and information structure. Iterate over the written summary multiple times to improve it, and re-read the reviews whenever necessary. 
    \item Please write your summary as if it were a review itself, e.g. 'This place is expensive' instead of 'Users thought this place was expensive'. Keep the length of the summary reasonably close to the average length of the reviews.
    \item Please try to write the summary using your own words instead of copying text directly from the reviews. Using the exact words from the reviews is allowed, but do not copy more than 5 consecutive words from a review . 
\end{itemize}

\subsection{Latent Codes Analysis}


\begin{table*}
 	\footnotesize 
 	\centering
    \begin{tabular}{ | >{\centering\arraybackslash} m{1.cm} | m{12cm} |}
    \hline
    mean $z$ & Bought this for my Kindle Fire HD and it works great. I have had no problems with it. I would recommend it to anyone looking for a good quality cable.\\ \hline
    $z_1$ &  Works fine with my Kindle Fire HD 8.9". The picture quality is very good, but it doesn't work as well as the picture. I'm not sure how long it will last, but i am very disappointed.\\ \hline
    $z_2$ & This is a great product. I bought it to use with my Kindle Fire HD and it works great. I would recommend it to anyone who is looking for a good quality cable for the price.\\ \hline
    $z_3$ & Good product, does what it is supposed to do. I would recommend it to anyone looking for a HDMI cable.\\ \hline \hline
    Rev 1 & Love this HDMI cable , but it only works with HD Kindle and not the HDX Kindle which makes me kinda crazy . I have both kinds of Kindles but the HDX is newer and I can 't get a cable for the new one . I guess my HD Kindle will be my Amazon Prime Kindle . It works great ! \\ \hline
    Rev 2 & I got a kindle for Christmas . I had no idea how to work one etc . I discovered you can stream movies to your tv and this is the exact cable for it . Works great and seems like its good quality . A bit long though. \\ \hline
    Rev 3 & this is great for watching movies from kindle to tv . Now the whole family can enjoy rather than one person at a time . Picture quality isn 't amazing , but it 's good . \\ \hline
    Rev 4 & I just received this wire in the mail , and it does not work in the slightest . I am very displeased with this product . \\ \hline
    Rev 5 & Works great ! ! Now I can watch Netflix on my TV with my Kindle Fire HD ... I love it and so will you ! \\ \hline
    Rev 6 & Works awesome . Great item for the price.Got it very quickly . Was as described in the ad.Exactly what I was looking for. \\ \hline
    Rev 7 & I plugged it into my Kindle fire HD and into the TV and works perfectly . Have had no problems with it ! \\ \hline
    Rev 8 & This is just what I was looking for to connect my Kindle Fire to view on our TV ! Great price too!\\ \hline
    \end{tabular}
    \caption{Amazon summaries of the full model with sampled and mean assignment to $z$. The assignment to $c$ was fixed, and was the mean value based on the approximate posterior $q_{\phi}(c|r_1,...,r_N)$.}
    \label{table:ama_latent_analysis_summs2}
\end{table*}

We performed a qualitative analysis of the latent variable $z$ to shed additional light on what it stores and sensitivity of the decoder with respect to its input. Specifically, we computed the mean value for the variable $c$ using the approximate posterior $q_{\phi}(c|r_1,...,r_N)$, and then sampled $z$ from the prior $p_{\theta}(z|c)$. 

First, we observed that the summaries produced using the mean of $z$ are more fluent. For example, in Table \ref{table:ama_latent_analysis_summs2}, the $z_1$ based summary states: \enquote{The picture quality is very good, but it doesn’t work aswell as the picture.}, where the second phrase could be rewritten in a more fluent matter.
Also, we found that mean based summaries contain less details that are partially or not supported by the reviews. For example, in the table, $z_1$ based summary mentions Kindle Fire HD 8.9', while the dimension is never mentioned in the reviews. 
Finally, different samples were observed to result in texts that contain different details about the reviews. For example, $z_1$ sample results in the summary that captures the picture quality, while $z_3$ that the item is good for its price. 
Overall, we observed that the latent variable $z$ stores content based information, that results in syntactically diverse texts, yet reflecting information about the same businesses or product.

\subsection{Repetitions}
We observed an increase in the amount of generated repetitions both in the reconstructed reviews and summaries when the $z$-related KL term is low and beam search is used. Intuitively, the initial input to the decoder becomes less informative, and it starts relying on learned local statistics to perform reconstruction. When the KLD vanishes to zero, the decoder essentially becomes a uncoditional language model, for which beam search was shown to lead to generation of repetitions \citep{holtzman2019curious}.






\begin{table*}[ht!]
 	\footnotesize 
 	\centering
    \begin{tabular}{ | >{\centering\arraybackslash} m{1.5cm} | m{12cm} |}
    \hline
    Ours & This place is the best Mexican restaurant i have ever been to. The food was delicious and the staff was very friendly and helpful. Our server was very attentive and made sure we were taken care of. We'll be back for sure. \\ \hline
    MeanSum & A little on the pricey side but I was pleasantly surprised. We went there for a late lunch and it was packed with a great atmosphere, food was delicious and the staff was super friendly. Very friendly staff. We had the enchiladas with a few extra veggies and they were delicious! Will be back for sure! \\ \hline
    LexRank & We will definitely be going back for more great food! Everything we had so far was great. The staff was great and so nice! Good food! Great atmosphere! \\ \hline
    Gold & This place is simply amazing! Its the best Mexican spot in town. Their tacos are delicious and full of flavor. They also have chips and salsa that is to die for! The salsa is just delectable!  It has a sweet, tangy flavor that you can't find anywhere else. I highly recommend! \\ \hline \hline
    Rev 1 &  Classic style Mexican food done nicely!  Yummy crispy cheese crisp with a limey margarita will will win my heart any day of the week!  The classic frozen with a chambord float is my favorite and they do it well here.The salad carbon was off the chain- served on a big platter and worked for me as 2 full dinners.\\ \hline
    Rev 2 & For delicious Mexican food in north Phoenix, try La Pinata. This was our visit here and we were so stunned by the speed in which our food was prepared that we were sure it was meant for another table. The food was hot and fresh and well within our budget. My husband got a beef chimichanga and I got bean and cheese burrito, which we both enjoyed. Chips and salsa arrived immediately; the salsa tastes sweeter than most and is equally flavorful. We will be back!  \\ \hline
    Rev 3 &  Good food! Great atmosphere! Great patio. Staff was super friendly and accommodating! We will definately return!\\ \hline
    Rev 4 & This place was very delicious! I got the ranchero burro and it was so good. The plate could feed at least two people. The staff was great and so nice! I also got the fried ice cream it was good. I would recommend this place to all my friends. \\ \hline
    Rev 5 & We arrive for the first time, greeted immediately with a smile and seated promptly. Our server was fantastic, he was funny and fast. Gave great suggestions on the menu and we both were very pleased with the food, flavors, speed and accuracy of our orders. We will definitely be going back for more great food! \\ \hline
    Rev 6 & Well was very disappointed to see out favorite ice cream parlor closed but delightfully surprised at how much we like this spot!!Service was FANTASTIC TOP notch!! Taco was great lots of cheese. Freshly deep fried shell not like SO MANY Phoenix mex restaurants use!  Enchilada was very good. My wife really enjoyed her chimichanga.  My moms chilli reanno was great too.  Everything we had so far was great.  We will return. Highly recommended. \\ \hline
    Rev 7 & I'm only on the salsa and it's just as fabulous as always. I love the new location and the decor is beautiful. Open 5 days and the place is standing room only. To the previous negative commentor, they are way took busy to fill an order for beans. Go across the street....you'll be angry lol. \\ \hline
    Rev 8 & I just tried to make a reservation for 15 people in March at 11 am on a Tuesday and was informed by a very rude female.  She said "we do not take reservations" and I asked if they would for 15 people and she said " I told you we don't take reservations" and hung up on me.  Is that the way you run a business?  Very poor customer service and I have no intentions of ever coming there or recommending it to my friends. \\ \hline
    \end{tabular}
    \caption{Yelp summaries produced by different models.}
    \label{table:yelp_example_summs2}
\end{table*}

\begin{table*}
 	\footnotesize 
 	\centering
    \begin{tabular}{ | >{\centering\arraybackslash} m{1.5cm} | m{12cm} |}
    \hline
    Ours & This place is the worst service I've ever had. The food was mediocre at best. The service was slow and the waiter was very rude. I would not recommend this place to anyone who wants to have a good time at this location. \\ \hline
    MeanSum  & I love the decor, but the food was mediocre. Service is slow and we had to ask for refills. They were not able to do anything and not even charge me for it.  It was a very disappointing experience and the service was not good at all. I had to ask for a salad for a few minutes and the waitress said he didn't know what he was talking about. All I can say is that the staff was nice and attentive. I would have given 5 stars if I could. \\ \hline
    LexRank & Food was just okay, server was just okay. The atmosphere was great, friendly server. It took a bit long to get a server to come over and then it took our server a while to get our bread and drinks. However there was complementary bread served.The Pizza I ordered was undercooked and had very little sauce.Macaroni Grill has unfortunately taken a dive. Went to dinner with 4 others and had another bad experience at the Macaroni Grill. \\ \hline
    Gold &  I'm really not a fan of Macaroni Grill, well, at least THIS Macaroni Grill.  The staff is slow and really doesn't seem to car about providing quality service.  It took well over 30 minutes to get my food and the place wasn't even packed with people.  I ordered pizza and it didn't taste right.  I think it wasn't fully cooked.  I won't be coming back.\\ \hline \hline
    Rev 1 & 10/22/2011 was the date of our visit.  Food was just okay, server was just okay.  The manager climbed up on the food prep counter to fix a light.  We felt like that was the most unsanitary thing anyone could do - he could have just come from the restroom for all we knew.  Needless to say, lackluster service, mediocre food and lack of concern for the cleanliness of the food prep area will guarantee we will NEVER return.\\ \hline
    Rev 2 &  We like the food and prices are reasonable. Our biggest complaint is the service. It took a bit long to get a server to come over and then it took our server a while to get our bread and drinks. They really need to develop a better sense of teamwork.  While waiting for things there were numerous servers standing around gabbing. It really gave us the impression of "Not my table." "Not my problem." Only other complaint is they need to get some rinse aid for the dishwasher. I had to dry our bread plates when the hostess gave them to us.\\ \hline
    Rev 3 & Not enough staff is on hand the two times I have been in to properly pay attention to paying customers. I agree that the portions have shrunk over the years, and the effort is no longer there. It is convenient to have nearby but not worth my time when other great restaurants are around. Wish I could rate it better but it's just not that good at all. \\ \hline
    Rev 4 & Went to dinner with 4 others and had another bad experience at the Macaroni Grill.  When will we ever learn?  The server was not only inattentive, but p o'd when we asked to be moved to another table.  When the food came it was at best, luke warm.  They had run out of one of our ordered dishes, but didn't inform us until 20 minutes after we had ordered.  Running out at 6:00 p.m.: Really? More delay and no apologies. There is no excuse for a cold meal and poor service.  We will not go back since the Grill seems not to care and there are plenty of other restaurants which do. \\ \hline
    Rev 5 & The service is kind and friendly. However there was complementary bread served.The Pizza I ordered was undercooked and had very little sauce.Macaroni Grill has unfortunately taken a dive.  Best to avoid the place or at the very least this location. \\ \hline
    Rev 6 & I know this is a chain, but Between this and Olive Garden, I would def pick this place. Service was great at this location and food not bad at all, although not excellent, I think it still deserves a good 4 stars \\ \hline
    Rev 7 &  I had a  2 for 1 \$9.00 express dinner coupon so we order up 2 dinners to go. The deal was 9 min or its free, it took 20, but since I was getting 2 meals for \$9.00 I did not make a fuss.  The actual pasta was fine and amount was fair but it  had maybe a 1/4 of a chicken breast. The chicken tasted like it came from Taco Bell, VERY processed. The sauce straight from a can. I have had much better frozen dinners. My husband and I used to like Macaroni Grill it sad too see its food go so down hill.\\ \hline
    Rev 8 & The atmosphere was great, friendly server. Although the food I think is served from frozen. I ordered mama trio. The two of three items were great. Plate came out hot, couldn't touch it. Went to eat lasagna and was ice cold in the center, nit even warm. The server apologized about it offered new one or reheat this one. I chose a new one to go. I saw her go tell manager. The manager didn't even come over and say anything. I was not even acknowledged on my way out and walked past 3 people. I will not be going back. Over priced for frozen food. \\ \hline
    \end{tabular}
    \caption{Yelp summaries produced by different models.}
    \label{table:yelp_example_summs4}
\end{table*}
\begin{table*}
 	\footnotesize 
 	\centering
    \begin{tabular}{ | >{\centering\arraybackslash} m{1.5cm} | m{12cm} |}
    \hline
    Ours & My wife and i have been here several times now and have never had a bad meal. The service is impeccable, and the food is delicious. We had the steak and lobster, which was delicious. I would highly recommend this place to anyone looking for a good meal. \\ \hline
    MeanSum & Our first time here, the restaurant is very clean and has a great ambiance. I had the filet mignon with a side of mashed potatoes.  They were both tasty and filling. I've had better at a chain restaurant, but this is a great place to go for a nice dinner or a snack.  Have eaten at the restaurant several times and have never had a bad meal here. \\ \hline
    LexRank & Had the filet... Really enjoyed my filet and slobster. In addition to excellent drinks, they offer free prime filet steak sandwiches. I have had their filet mignon which is pretty good, calamari which is ok, scallops which aren't really my thing, sour dough bread which was fantastic, amazing stuffed mushrooms. Very good steak house. \\ \hline
    Gold &   The steak is the must have dish at this restaurant. One small problem with the steak is that you want to order it cooked less than you would at a normal restaurant. They have the habit of going a bit over on the steak. The drinks are excellent and the stuffed mushrooms as appetizers were amazing. This is a classy place that is also romantic. The staff pays good attention to you here. \\ \hline \hline
    Rev 1 & The ambiance is relaxing, yet refined. The service is always good. The steak was good, although not cooked to the correct temperature which is surprising for a steakhouse. I would recommend ordering for a lesser cook than what you normally order. I typically order medium, but at donovan's would get medium rare. The side dish menu was somewhat limited, but we chose the creamed spinach and asparagus, both were good. Of course, you have to try the creme brulee - Yum! \\ \hline
    Rev 2 & Hadn't been there in several years and after this visit I remember why, I don't like onions or shallots in my macaroni and cheese. The food is good but not worth the price just a very disappointing experience and I probably won't go back \\ \hline
    Rev 3 & My wife and I come here every year for our anniversary (literally every year we have been married). The service is exceptional and the food quality is top-notch. Furthermore, the happy hour is one of the best in the Valley. In addition to excellent drinks, they offer free prime filet steak sandwiches. I highly recommend this place for celebrations or a nice dinner out. \\ \hline
    Rev 4 & I get to go here about once a month for educational dinners.  I have never paid so don't ask about pricing.  I have had their filet mignon which is pretty good, calamari which is ok, scallops which aren't really my thing, sour dough bread which was fantastic, amazing stuffed mushrooms.  The vegetables are perfectly cooked and the mashed potatoes are great.  At the end we get the chocolate mousse cake that really ends the night well.  I have enjoyed every meal I have eaten there. \\ \hline
    Rev 5 & Very good steak house.  Steaks are high quality and the service was very professional.  Attentive, but not hovering.  Classic menus and atmosphere for this kind of restaurant. No surprises. A solid option, but not a clear favorite compared to other restaurants in this category. \\ \hline
    Rev 6 & Had a wonderful experience here last night for restaurant week. Had the filet... Which was amazing and cooked perfectly with their yummy mashed potatoes and veggies. The bottle of red wine they offered for an additional \$20 paired perfectly with the dinner. The staff were extremely friendly and attentive. Can't wait to go back! \\ \hline
    Rev 7 & The seafood tower must change in selection of seafood, which is good, which is also why mine last night was so fresh fresh delicious.  Its good to know that you can get top rate seafood in Phoenix.  Bacon wrapped scallops were very good, and I sacrificied a full steak (opting for the filet medallion) to try the scallops.  I asked for medium rare steak, but maybe shouldve asked for rare...my cousin had the ribeye and could not have been any happier than he was :)  yum for fancy steak houses.  Its an ultra romantic place to, fyi.the wait staff is very attentive. \\ \hline
    Rev 8 & Donovans, how can you go wrong.  Had some guests in town and some fantastic steaks paired with some great cabernets.  Really enjoyed my filet and lobster. \\ \hline
    \end{tabular}
    \caption{Yelp summaries produced by different models.}
    \label{table:yelp_example_summs3}
\end{table*}


\begin{table*}[h]
 	\footnotesize 
 	\centering
    \begin{tabular}{ | >{\centering\arraybackslash} m{1.5cm} | m{12cm} |}
    \hline
    Ours & I love this tank. It fits well and is comfortable to wear. I wish it was a little bit longer, but I'm sure it will shrink after washing. I would recommend this to anyone.\\ \hline
    MeanSum & I normally wear a large so it was not what I expected. It's a bit large but I think it's a good thing. I'm 5 '4 "and the waist fits well. I'm 5 '7 and this is a bit big.\\ \hline
    LexRank & I'm 5 '4 'and this tank fits like a normal tank top, not any longer. The only reason I'm rating this at two stars is because it is listed as a 'long' tank top and the photo even shows it going well past the models hips, however I'm short and the tank top is just a normal length. I bought this tank to wear under shirts when it is colder out. I was trying to find a tank that would cover past my hips, so I could wear it with leggings.\\ \hline
    Gold & Great tank top to wear under my other shirts as I liking layering and the material has a good feel. There was a good choice of colors to pick from. Although, the top is a thin material I don't mind since I wear it under something else.\\ \hline \hline
    Rev 1 & The description say it long... NOT so it is average. That's why I purchased it because it said it was long. This is a basic tank.I washed it and it didn't warp but did shrink a little. Nothing to brag about. \\ \hline
    Rev 2 & I'm 5 '4 'and this tank fits like a normal tank top, not any longer. I was trying to find a tank that would cover past my hips, so I could wear it with leggings. Don't order if you're expecting tunic length.\\ \hline
    Rev 3 & This shirt is OK if you are layering for sure. It is THIN and runs SMALL. I usually wear a small and read the reviews and ordered a Medium. It fits tight and is NOT long like in the picture. Glad I only purchased one. \\ \hline
    Rev 4 & The tank fit very well and was comfortbale to wear. The material was thinner than I expected, and I felt it was probably a little over priced. I've bought much higher quality tanks for \$5 at a local store. \\ \hline
    Rev 5 & The only reason I'm rating this at two stars is because it is listed as a 'long' tank top and the photo even shows it going well past the models hips, however I'm short and the tank top is just a normal length. \\ \hline
    Rev 6 & I usually get them someplace out but they no longer carry them. I thought I would give these a try. I received them fast, although I did order a brown and got a black (which I also needed a black anyway). They were a lot thinner than I like but they are okay.\\ \hline
    Rev 7 & Every women should own one in every color. They wash well perfect under everything. Perfect alone. As I write I'm waiting on another of the same style to arrive. Just feels quality I don't know how else to explain it, but I'm sure you get it ladies!\\ \hline
    Rev 8 & I bought this tank to wear under shirts when it is colder out. I bought one in white and one in an aqua blue color. They are long enough that the color peeks out from under my tops. Looks cute. I do wish that the neck line was a bit higher cut to provide more modest coverage of my chest.\\ \hline
    \end{tabular}
    \caption{Amazon summaries produced by different models.}
    \label{table:ama_example_summs1}
\end{table*}

\begin{table*}[h]
 	\footnotesize 
 	\centering
    \begin{tabular}{ | >{\centering\arraybackslash} m{1.5cm} | m{12cm} |}
    \hline
    Ours &  This is the best acupressure mat I have ever used. I use it for my back pain and it helps to relieve my back pain. I have used it for several months now and it seems to work well. I would recommend it to anyone.\\ \hline
    MeanSum &  I have used this for years and it works great. I have trouble with my knee pain, but it does help me to get the best of my feet. I have had no problems with this product. I have had many compliments on it and is still in great shape. \\ \hline
    LexRank & I ordered this acupressure mat to see if it would help relieve my back pain and at first it seemed like it wasn't doing much, but once you use it for a second or third time you can feel the pain relief and it also helps you relax. its great to lay on to relax you after a long day at work. I really like the Acupressure Mat. I usually toss and turn a lot when I sleep, now I use this before I go to bed and it helps relax my body so that I can sleep more sound without all the tossing and turning. \\ \hline
    Gold & These acupressure mats are used to increase circulation and reduce body aches and pains and are most effective when you can fully relax. Consistence is key to receive the full, relaxing benefits of the product. However, if you are using this product after surgery it is responsible to always consult with your physician to ensure it is right for your situation. \\ \hline \hline
    Rev 1 & Always consult with your doctor before purchasing any circulation product after surgery. I had ankle surgery and this product is useful for blood circulation in the foot. This increase in circulation has assisted with my ability to feel comfortable stepping down on the foot (only after doc said wait bearing was okay). I use it sitting down barefoot. \\ \hline
    Rev 2 & I really like the Acupressure Mat. I usually toss and turn a lot when I sleep, now I use this before I go to bed and it helps relax my body so that I can sleep more sound without all the tossing and turning.\\ \hline
    Rev 3 & I used the mat the first night after it arrived and every-other night since. After 2 ten minute sessions, I am sold. I have slept much better at night - I think it puts me in a more relaxed state, making it easier to fall asleep. A rather inexpensive option to relieving tension in my neck, upper back and shoulders. \\ \hline
    Rev 4 & This is the best thing! you can use socks if your feet are tender to walk on it or bare foot if you can take it. I use it every morning to walk across to jump start my body. when I think about it I will lay on it, it feels wonderful. \\ \hline
    Rev 5 & I love these spike mats and have recommended them to everyone that has had any kind of body ache. its great to lay on to relax you after a long day at work. Helps with pain in my back and pain in my legs. Its not a cure, but it sure helps with the healing process. \\ \hline
    Rev 6 & I wish I hadn't purchased this item. I just can't get use to it, it's not comfortable. I have not seen any benefits from using it but that could be because I don't relax or use it for long enough. \\ \hline
    Rev 7 & I run an alternative health center and use Acupressure pin mats from different sources to treat my patients, but this product is the patients choice, they are asking allways for this mat against other brands so I changed all of them for Britta, moreover the S \& H was outstanding and really fast.\\ \hline
    Rev 8 & I ordered this acupressure mat to see if it would help relieve my back pain and at first it seemed like it wasn't doing much, but once you use it for a second or third time you can feel the pain relief and it also helps you relax. I use it almost everyday now and it really helps. I recommed this product and this seller. \\ \hline
    \end{tabular}
    \caption{Amazon summaries produced by different models.}
    \label{table:ama_example_summs3}
\end{table*}


\end{document}